\documentclass[]{bytedance_seed}

\usepackage[toc,page,header]{appendix}

\usepackage{minitoc}
\usepackage[utf8]{inputenc}
\usepackage{multirow} 
\usepackage[table]{xcolor}
\usepackage{cleveref}
\usepackage{amssymb}
\usepackage{graphicx}

\title{SIMART: Decomposing Monolithic Meshes into Sim-ready Articulated Assets via MLLM}

\author[1,2,*]{Chuanrui Zhang}
\author[1,\#,\dagger]{Minghan Qin}
\author[1]{Yuang Wang}
\author[1]{Baifeng Xie} 
\author[1]{Hang Li}
\author[2, \dagger]{Ziwei Wang}

\affiliation[1]{ByteDance Seed}
\affiliation[2]{NTU}

\contribution[*]{Work done at ByteDance Seed}
\contribution[\#]{Project Lead}
\contribution[\dagger]{Corresponding authors}

\abstract{
High-quality articulated 3D assets are indispensable for embodied AI and physical simulation, yet 3D generation still focuses on static meshes, leaving a gap in "sim-ready" interactive objects. Most recent articulated object creation methods rely on multi-stage pipelines that accumulate errors across decoupled modules. Alternatively, unified MLLMs offer a single-stage path to joint static asset understanding and sim-ready asset generation. However dense voxel-based 3D tokenization yields long 3D token sequences and high memory overhead, limiting scalability to complex articulated objects. To address this, we propose SIMART, a unified MLLM framework that jointly performs part-level decomposition and kinematic prediction. By introducing a Sparse 3D VQ-VAE, SIMART reduces token counts by 70\% vs. dense voxel tokens, enabling high-fidelity multi-part assemblies. SIMART achieves state-of-the-art performance on PartNet-Mobility and in-the-wild AIGC datasets, and enables physics-based robotic simulation.
}

\checkdata[Project Page]{\url{https://simart-mllm.github.io}}

\correspondence{Minghan Qin at \email{minghan@bytedance.com}}

\begin{document}
\maketitle


\section{Introduction}

\begin{figure}
  \includegraphics[width=\textwidth]{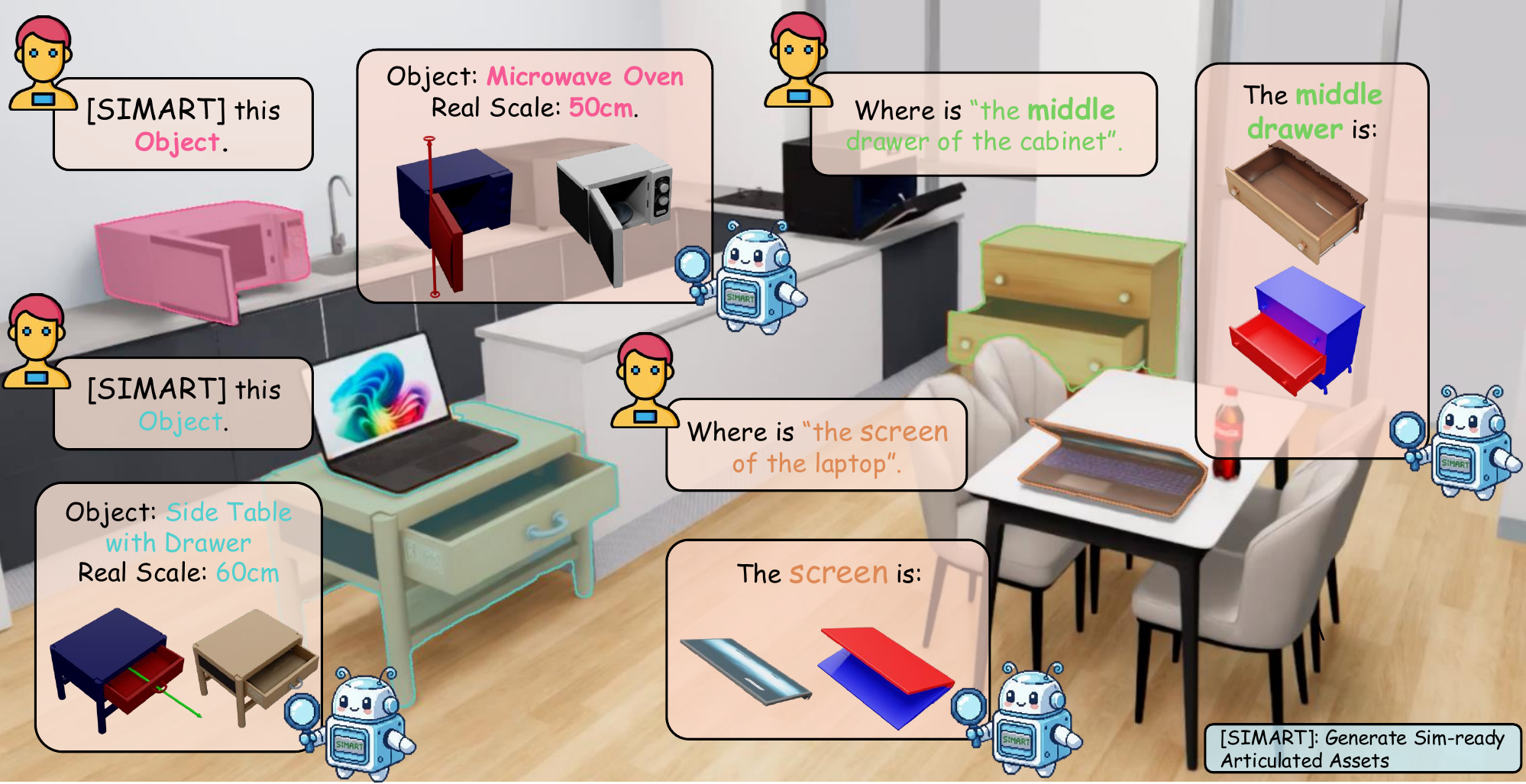}
  \caption{SIMART leverages the multimodal reasoning of MLLMs to unify URDF generation and semantic part grounding, transforming static 3D meshes into functional, simulation-ready articulated assets.}
  \label{fig:teaser}
\end{figure}

Simulation-ready articulated assets have garnered significant attention recently as physical properties and kinematic information are essential for physics-based animation and robotic interactive simulation~\cite{xiang2020sapien, liu2025survey}. 
However, a vast majority of existing 3D assets remain unarticulated~\cite{Mo_2019_CVPR}. 
Furthermore, the manual creation of such assets is prohibitively labor-intensive, underscoring the critical need for robust, automated methods for sim-ready asset generation.

While directly regressing URDF models from images is intuitive \cite{chen2024urdformer,le2024articulate}, it often compromises geometric fidelity, yielding coarse outputs unsuitable for high-quality simulation. To circumvent this, prior methods for generating articulated objects often adopt multi-stage pipelines---decoupling the process into part decomposition, joint parameter inference, and post-hoc assembly~\cite{paschalidou2020learning,xu2022universal}. Part decomposition is frequently not articulation-aware: prompting 2D vision(-language) models transfers unreliably to 3D boundaries, while 3D-native segmentation methods (e.g., PartField \cite{liu2025partfield} and P3SAM \cite{ma2025p3}) primarily optimize for surface-level consistency and can miss mechanically meaningful link boundaries, producing parts that look plausible yet violate kinematic affordances. Joint estimation further amplifies this fragility. Whether predicted from 2D cues or optimized from imperfect geometry is sensitive to mesh artifacts and restrictive priors, so the inferred joints often become incompatible with the recovered part geometry, yielding physically invalid articulation.

Meanwhile, recent 3D generative models \cite{yang2024hunyuan3d,seed2025seed3d,li2024advances,tang2023dreamgaussian} can synthesize high-quality static assets, but these outputs are typically monolithic, non-decomposed meshes without any kinematic or physical metadata. 
This motivates a unified MLLM paradigm that \emph{understands} such an initial 3D asset and directly \emph{generates} per-part geometry (as tokens) together with a structured URDF specification (links, joints axes, and limits). However, existing 3D-native MLLM attempts \cite{ye2025shapellm,cao2025physx,cao2025physx-a} are constrained by inefficient 3D tokenization: dense volumetric encodings \cite{ye2025shapellm,xiang2025structured} waste most tokens on empty space, leading to prohibitive context length and memory overhead even for understanding, let alone part-level generation. A central technical challenge is thus to develop an efficient, sparse 3D representation that simultaneously supports MLLM-based understanding and scalable, high-fidelity generation.

To address these limitations, we propose \textbf{SIMART}, a unified multimodal architecture that integrates 3D geometric understanding and generation. 
This integration enables the model to jointly perform part-level mesh decomposition and precise kinematic parameter prediction simultaneously. 
To overcome the computational bottlenecks of dense voxel representations, we introduce a Sparse 3D VQ-VAE that selectively encodes occupied surface voxels. This refinement reduces token counts by 70\%, effectively mitigating memory exhaustion and enabling the detailed articulation modeling of intricate assemblies that were previously computationally prohibitive.
To evaluate our approach, we curate SIMART-Bench, a high-quality benchmark consolidating assets from PartNet-Mobility and diverse generative sources, refined with expert manual annotations to ensure articulation accuracy. Experimental results demonstrate that SIMART significantly outperforms existing state-of-the-art models, exhibiting superior generalization in transforming diverse static meshes into simulation-ready assets.
Furthermore, we demonstrate the downstream utility of our generated assets in physics-based simulation and VR/AR applications.

The contributions of this work are summarized as follows:

\begin{itemize}
\item We propose a novel, unified MLLM framework designed to directly perceive and generate kinematically-aware meshes along with their underlying kinematic logic.
\item We introduce a Sparse 3D VQ-VAE representation that reduces token redundancy by $70\%$, thereby facilitating efficient MLLM processing of complex 3D meshes.
\item We propose a high-fidelity articulation benchmark and demonstrate state-of-the-art performance in part decomposition and joint parameter estimation.

\end{itemize}
\section{Related Work}

\subsection{Articulated Object Reconstruction and Generation}
Reconstruction-based methods \cite{jiang2022ditto, liu2023paris, liu2025building,deng2024articulate,weng2024neural,guo2025articulatedgs,gao2025partrm,xia2025drawer,zhang2025iaao} leverage neural representations like Neural Radiance Fields (NeRF) \cite{mildenhall2021nerf} or 3D Gaussian Splatting (3DGS) \cite{kerbl20233d} to recover high-fidelity geometries. For instance, ArtGS \cite{liu2025building} and ArticulatedGS \cite{guo2025articulatedgs} incorporate motion constraints to extract kinematic structures from observed states. However, these methods typically require multi-view supervision across different articulation stages (e.g., images of a cabinet both open and closed). Such high-quality, multi-state visual inputs are often difficult to obtain in the wild, leading to poor generalization when faced with incomplete observations or sparse viewpoints. Generative-based methods \cite{liu2024cage,lei2023nap,qiu2025articulate,gao2025meshart,liu2024singapo,chen2024urdformer,lian2025infinite,wu2025dipo}, such as CAGE \cite{liu2024cage} and SINGAPO \cite{liu2024singapo}, attempt to mitigate these constraints by learning category-level priors through Diffusion Models or part-based slots. Nevertheless, these frameworks are hindered by the acute scarcity and limited diversity of articulated 3D datasets compared to rigid objects. Consequently, these models are prone to overfitting, often failing to produce structurally sound or novel articulations for uncommon object categories.

\subsection{MLLM for articulation}

Recent frameworks, such as Articulate-Anything~\cite{le2024articulate} and Articulate AnyMesh~\cite{qiu2025articulate}, leverage the visual reasoning capabilities of MLLMs to infer motion structures from rendered images. However, these methods lack an integrated 3D geometric understanding and generation pathway, as they rely exclusively on 2D visual inputs.
PhysX-Anything~\cite{cao2025physx-a} leverages MLLMs to generate 3D voxels; however, it still struggles with capturing fine-grained spatial information due to the heavy computational overhead of dense voxel tokens.
The integration of MLLMs with 3D perception \cite{ahmed2025kestrel,fang2025meshllm,li2025urdf,ye2025shapellm} has evolved from general-purpose shape captioning to specialized kinematic reasoning. 
However, these view-dependent methods often lack direct geometric grounding, leading to physically inconsistent joint estimations. To address this, 3D-native models via MLLM \cite{li2025urdf} attempt to encode volumetric features for structural parsing. 
Despite their promise, these approaches frequently rely on dense volumetric tokenization, which introduces massive computational redundancy by encoding empty space. 
This $O(N^3)$ complexity not only triggers memory-exhaustion on complex meshes but also necessitates heavy downsampling that compromises the geometric fidelity required for precise axis localization. 
Furthermore, the inherent task interference in end-to-end generative-articulation paradigms often leads to suboptimal structural accuracy.

\begin{figure}[t]
  \centering
  \includegraphics[width=\linewidth]{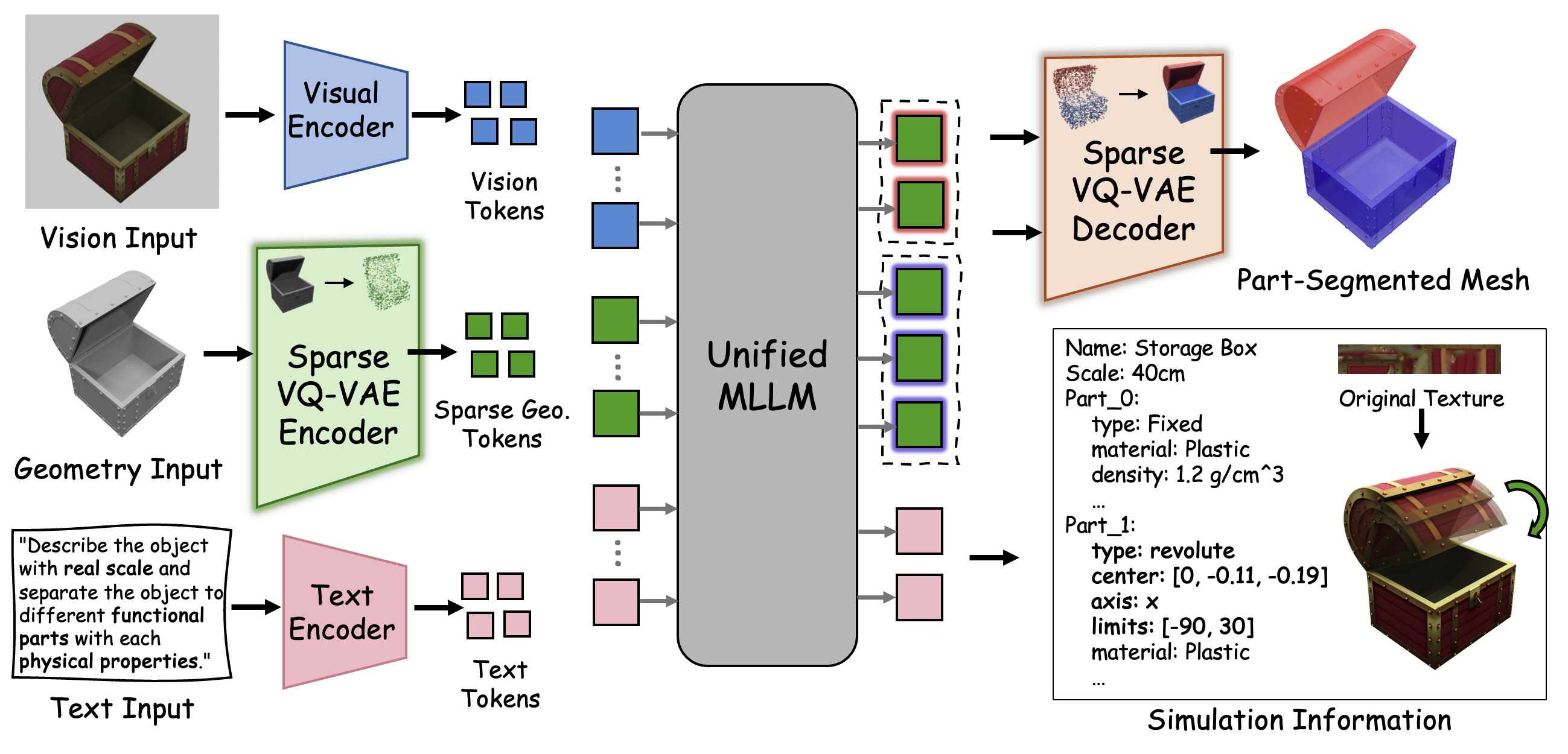}
    \caption{
    The pipeline of our SIMART. 
    The framework first encodes 3D geometry into a compact representation using the Sparse 3D VQ-VAE to minimize token redundancy while preserving critical surface details. 
    These geometric tokens are then fused with visual and textual inputs through a unified MLLM backbone to perform part grounding and joint parameter estimation. 
    The final output consists of structured URDF metadata and decomposed segments, enabling deployment into physics-based simulators and interactive robotic environments.
}
  \label{fig:pipeline}
\end{figure}

\subsection{3D Part Understanding}

Current 3D part understanding paradigms primarily oscillate between geometric precision and semantic flexibility. 
To achieve broad semantic coverage, recent 2D-to-3D lifting methods, such as PartField \cite{liu2025partfield} and P3SAM \cite{ma2025p3}, project foundational priors from large-scale models~\cite{kirillov2023segment, radford2021learning, oquab2023dinov2, simeoni2025dinov3} into 3D representations.
While effective for open-vocabulary recognition, these approaches often suffer from cross-view inconsistencies and "blurry" boundaries, failing to provide the structural rigor necessary for precise kinematic joint estimation. 
Concurrently, Gaussian-based \cite{kerbl20233d} architectures have evolved beyond pure appearance modeling to integrate semantic descriptors \cite{qin2024langsplat,li2025langsplatv2,li2024langsurf,chen2025slgaussian} and part-level decomposition with physical reasoning  \cite{xie2024physgaussian,abou2024physically,gao2025mani,lou2025robo,liu2025building}, enabling a more holistic understanding of volumetric reconstructions. 
However, these approaches remain observation-dependent, requiring dense temporal sequences or predefined kinematic templates to anchor dynamics. 
\section{Approach}

\label{sec:method}

In this section, we first formulate the task of transforming raw geometric observations into functional assets, and then elaborate on how our proposed SIMART framework addresses this challenge. \Cref{fig:pipeline} illustrates the overall pipeline.

\subsection{Problem Formulation}
Given a set of multimodal inputs $\mathcal{I} = \{I_{vis}, G_{geo}, T_{txt}\}$, where $I_{vis}$ represents the visual observations, $G_{geo}$ denotes the raw geometry data, and $T_{txt}$ is the language instruction describing the desired task, our objective is to generate a simulation-ready asset $\mathcal{A}$. 
Formally, the output asset $\mathcal{A}$ is defined by the tuple $(\mathcal{M}_{seg}, \mathcal{P}_{sim})$. Here, $\mathcal{M}_{seg} = \{m_1, m_2, \dots, m_n\}$ constitutes a set of part-segmented meshes representing decomposed functional components.
The term $\mathcal{P}_{sim}$ denotes a comprehensive set of simulation metadata that characterizes kinematic parameters such as joint types, axes, and limits, as well as dynamic physical properties including global scale, surface friction, and material density.

\subsection{Unified MLLM}
\label{sec:llm}

We adopt the Qwen3-VL~\cite{Qwen3-VL} architecture as our MLLM backbone, leveraging its powerful large-scale image-text pre-training and its emergent capability for physical world understanding. 
Unlike traditional 3D-specific models, this architecture enables our system to reason about abstract physical attributes, such as material properties and potential kinematic structures, by drawing upon a vast corpus of multimodal knowledge.

\noindent\textbf{Input.} 
The SIMART framework processes a heterogeneous sequence comprising vision, geometry, and text tokens, mapped into a unified latent space for joint reasoning. 
For the vision modality, an RGB image $I_{vis}$ is processed through a Vision Transformer (ViT) encoder to extract visual features $F_{vis} \in \mathbb{R}^{N_v \times D}$ that capture the object's semantic context.
To represent geometric features, the raw input mesh $G_{geo}$ is first discretized into a high-resolution voxel grid. 
These voxels are subsequently processed by the 3D-Unet encoder of our Sparse 3D VQ-VAE and quantized into discrete indices from a learned codebook, resulting in a sparse set of geometric tokens $F_{geo} \in \mathbb{R}^{N_g \times D}$. 
Concurrently, the text instruction $T_{txt}$ is embedded into tokens $F_{txt} \in \mathbb{R}^{N_t \times D}$ to steer the model toward specific task objectives. 
These instructions are designed to modulate the generation of fine-grained part grounding markers and structured URDF metadata, including kinematic hierarchies and physical properties. The final multimodal sequence, with a total length of $L = N_v + N_g + N_t$, is formed by concatenating these modality-specific features before being fed into the Transformer layers of the MLLM.


\noindent\textbf{Output.} 
The MLLM is optimized to generate a hybrid output sequence that satisfies both geometric constraints and symbolic structural requirements. 
Detailed specifications of the output format are provided in the Appendix.

\subsection{Sparse 3D VQ-VAE}
\label{sec:vqvae}

\begin{figure}[t]
  \centering
  \includegraphics[width=0.9\linewidth]{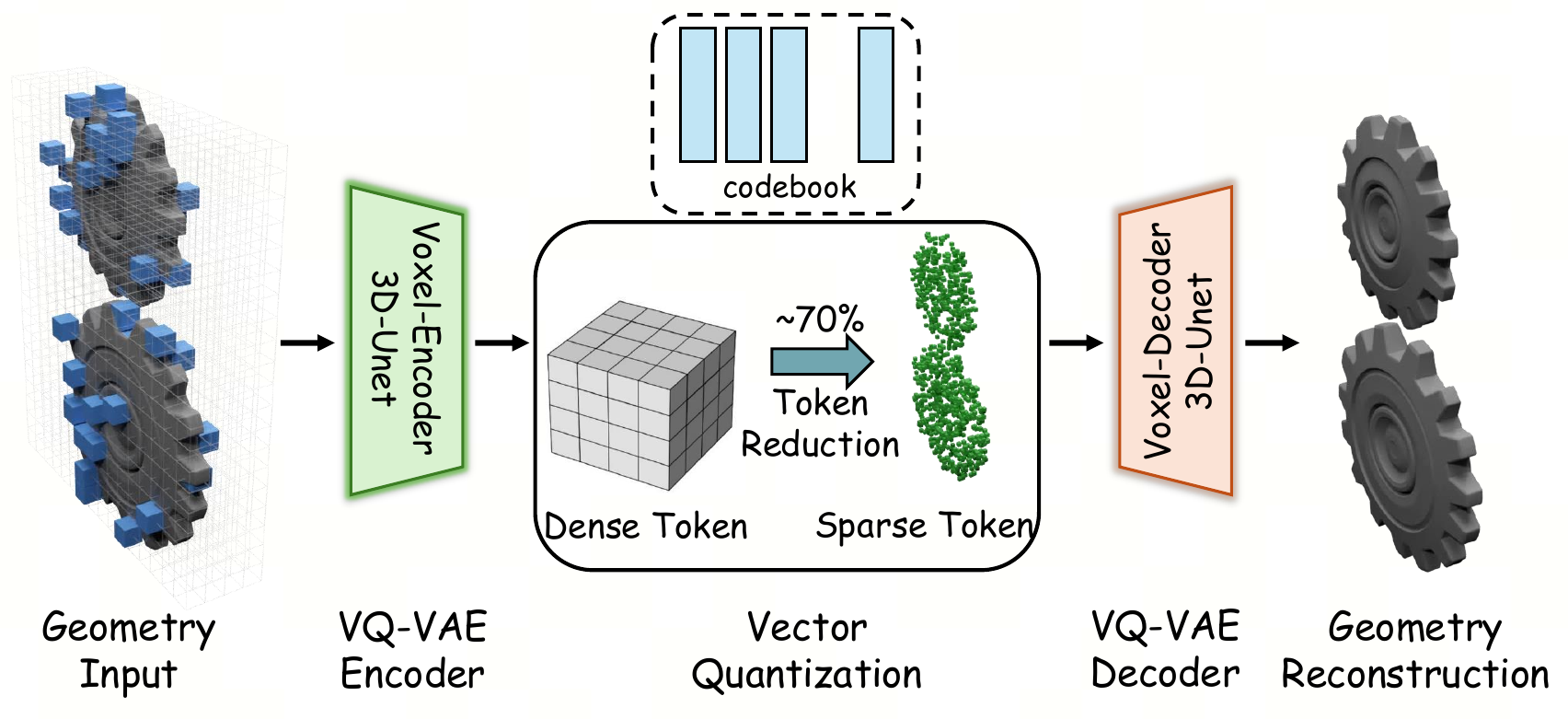}
  \caption{
  Architectural overview of the Sparse 3D VQ-VAE for high-fidelity geometric encoding. 
  The pipeline employs a 3D-UNet voxel encoder to map geometric inputs into a discrete latent space through vector quantization with a specialized codebook. 
  }
  \label{fig:vqvae}
\end{figure}

To incorporate complex 3D geometry into the MLLM framework, we adopt a voxel-based representation that achieves an optimal trade-off between spatial fidelity and computational efficiency, as illustrated in \Cref{fig:vqvae}.
Drawing inspiration from ShapeLLM-Omni~\cite{ye2025shapellm}, our architecture employs a 3D-Unet as the voxel encoder to map the origin $64^3$ grid into a compact latent feature grid $Z \in \mathbb{R}^{16 \times 16 \times 16 \times C}$, where $C$ denotes the feature dimension.
While traditional VQ-VAE models process every latent position regardless of occupancy, our method leverages the inherent sparsity of 3D data. 
We identify unoccupied voxels during the encoding process and assign them a specialized zero token ($\mathbf{e}_{zero}$) from our codebook $\mathcal{C}$. 
Only features corresponding to occupied geometric regions are passed to the vector quantization stage to find the nearest neighbor in the codebook. 
Formally, for each latent feature $z_i$ at index $i$, the quantized representation $\hat{z}_i$ is determined as:
\begin{equation}
    \hat{z}_i = 
    \begin{cases} 
    \mathbf{e}_{zero}, & \text{if Voxel } i \text{ is unoccupied} \\
    \arg\min_{\mathbf{e}_j \in \mathcal{C} \setminus \{\mathbf{e}_{zero}\}} \|z_i - \mathbf{e}_j\|_2, & \text{otherwise}
    \end{cases}
\end{equation}
This strategy allows the model to effectively bypass empty space, leading to a significant reduction in the number of informative tokens—approximately $70\%$ in typical scenarios—that the subsequent MLLM must process.

While dense representations rely on consistent sequence lengths to maintain a fixed mapping between indices and 3D coordinates, our sparse tokens require explicit localization to preserve the object's structural topology within the MLLM. 
Specifically, each occupied voxel is serialized into a triplet of atomic tokens in the format: $\langle \text{voxel} \rangle \ [xyz] \ [K]$. Here, $\langle \text{voxel} \rangle$ serves as a specialized start-of-voxel identifier, while $[K] \in [0, 4095]$ represents the discrete geometry index retrieved from the codebook $\mathcal{C}$. The spatial location is explicitly encoded by the coordinate token $[xyz]$, which is computed using a linearized index mapping $xyz = 64x + 8y + z$, where $x, y, z \in [0, 7]$ represent the discrete coordinates in a $8 \times 8 \times 8$ grid. 
This coordinate-aware tokenization allows the MLLM to perform fine-grained geometric reasoning over a variable-length sequence. 
For reconstruction, the sequence of discrete tokens is fed into a symmetric 3D-Unet architecture to recover the original $64^3$ geometry from the quantized latent space. 
By representing the 3D asset through this sparse voxel tokens, we provide the MLLM with a highly compressed yet structurally rich geometric foundation for articulation reasoning.

\subsection{Simulator-ready Assets Process}
\label{sec:process}

The final stage of our framework involves the synthesis of a functional, simulation-ready asset $\mathcal{A}$ by integrating reconstructed geometric components with structured kinematic metadata. 
Utilizing the Sparse 3D VQ-VAE decoder, the part-specific voxel tokens generated by the MLLM are decoded into sparse point clouds $S_p$. 
To map these discrete seeds onto the high-fidelity input mesh $G_{geo}$ and achieve precise part decomposition, we employ a robust graph-based surface segmentation algorithm. 
Specifically, we initialize a vertex-wise probability distribution across the mesh manifold, where the initial probability of a vertex $v$ belonging to a functional part $p$ is defined by a Gaussian kernel:
\begin{equation}
    P(v, p) \propto \exp\left(-\frac{d(v, S_p)^2}{2\sigma^2}\right)
\end{equation}
where $d(v, S_p)$ denotes the distance to the nearest seed of part $p$, and $\sigma$ is a scale hyperparameter relative to the mesh bounding box. 
To ensure boundary coherence, we apply an iterative graph-smoothing operator across the mesh adjacency matrix, assigning final face labels $\mathcal{M}_{seg}$ via majority voting.
Subsequently, the original texture of the input mesh is preserved and adopted as the final texture output.

Concurrently, SIMART directly generates a structured URDF specification that defines the asset's kinematic and dynamic logic. This specification encapsulates the kinematic chain (parent-child hierarchies and joint configurations) along with essential physical attributes, including joint limits, material density, and surface friction. By coupling the segmented sub-meshes with this articulated blueprint, we produce a simulation-ready asset capable of accurate inertial modeling and physical interaction.

\begin{table}[t]
  \centering
  \caption{Quantitative comparison of articulation accuracy and geometric fidelity. We evaluate performance across In-Domain items from PhysXNet and AI-generated objects to demonstrate the superior generalization of SIMART. \textbf{Bold} indicates the best performance, and \underline{underlined} indicates the second-best.}
  \label{tab:method_comparison}
  \resizebox{\columnwidth}{!}{
  \begin{tabular}{l|ccccc|ccccc}
    \toprule
    \multirow{2}{*}{Method} & \multicolumn{5}{c|}{ID Items} & \multicolumn{5}{c}{AI-generated Items} \\
    \cmidrule(lr){2-6} \cmidrule(lr){7-11}
    & Type $\uparrow$ & Axis $\downarrow$ & Origin $\downarrow$ & IOU $\uparrow$ & CD $\downarrow$ & Type $\uparrow$ & Axis $\downarrow$ & Origin $\downarrow$ & IOU $\uparrow$ & CD $\downarrow$ \\
    \midrule
    Urdformer~\cite{chen2024urdformer}           & 0.496 & 0.585 & 0.610 & 0.002 & 0.624 & 0.544 & 0.557 & 0.476 & 0.016 & 0.650 \\
    Articulate-Anything~\cite{le2024articulate} & 0.891 & 0.315 & 0.174 & 0.202 & 0.239 & 0.765 & 0.243 & 0.232 & 0.069 & 0.244 \\
    Physx-Anything~\cite{cao2025physx-a}      & 0.686 & 0.312 & 0.322 & 0.128 & 0.278 & 0.658 & 0.481 & 0.324 & 0.100 & 0.334  \\
    Particulate~\cite{li2025particulate}      & \underline{0.822} & \underline{0.208} & \underline{0.204} & \underline{0.643} & \underline{0.140} & \underline{0.817} & \underline{0.166} & \underline{0.168} & \underline{0.618} & \underline{0.106}  \\
    \midrule
    \rowcolor[gray]{0.9} \textbf{SIMART (ours)} & \textbf{0.928} & \textbf{0.080} & \textbf{0.111} & \textbf{0.690} & \textbf{0.087} & \textbf{0.831} & \textbf{0.136} & \textbf{0.145} & \textbf{0.777} & \textbf{0.079} \\
    \bottomrule
  \end{tabular}
}
\end{table}

\begin{figure*}[t]
  \centering
  \includegraphics[width=\linewidth]{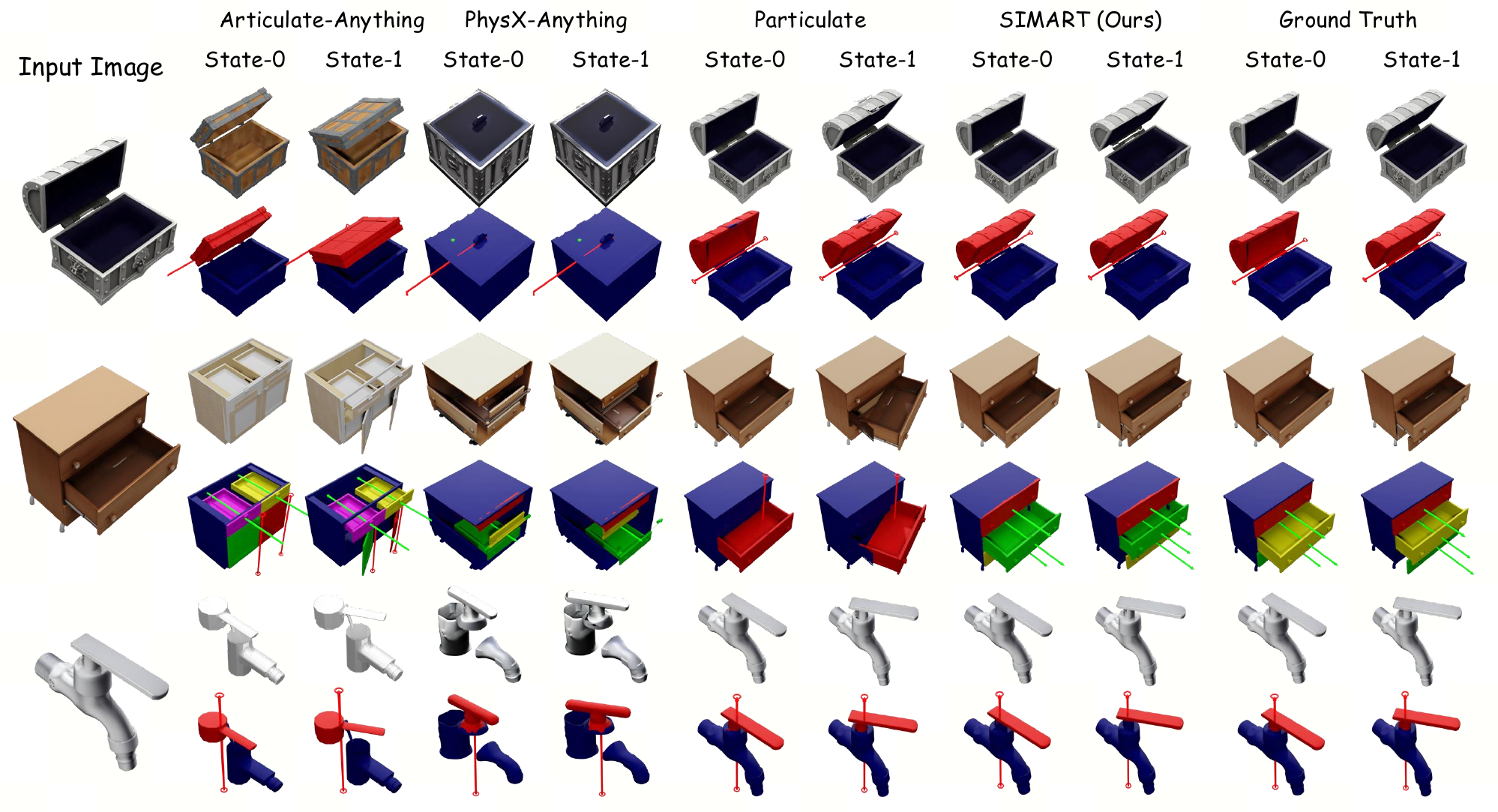}
  \caption{Qualitative comparison of articulated asset generation across different methods. Each object is visualized in two motion states to demonstrate kinematic accuracy and geometric fidelity. While existing generative baselines often produce simplified or misaligned meshes, SIMART achieves precise part-level segmentation and superior structural consistency, providing high-fidelity assets that closely match the ground-truth configurations.}
  \label{fig:m_cp}
\end{figure*}
\section{Experiments}

\label{sec:exp_setup}

\noindent \textbf{Dataset.} 
For the MLLM instruction tuning phase, we compile a dataset of 39,600 3D objects from PhysXNet~\cite{cao2025physx} and PartNet-Mobility~\cite{Mo_2019_CVPR}, which includes 5,600 articulated models along with 34,000 static objects intended to improve general shape comprehension. 
For data augmentation, we render each articulated model in 20 diverse kinematic states, effectively treating each state as an individual training instance.
Based on this collection, we synthesize two large-scale instruction-following datasets: a URDF generation set and a part grounding set, each containing 960k QA pairs. 
To rigorously evaluate generalization, we introduce SIMART-Bench, a high-fidelity benchmark addressing the limitations of existing datasets. While PartNet-Mobility is a standard resource, its data distribution is relatively homogeneous, with minimal geometric variance within categories. To overcome this lack of diversity, SIMART-Bench consolidates In-Domain assets from PartNet-Mobility with Out-of-Distribution (OOD) objects synthesized via AIGC (e.g., Hunyuan3d-V3.1~\cite{lai2025hunyuan3d}). This integration introduces diverse topologies that better challenge an algorithm's robustness beyond standard benchmarks.

\noindent \textbf{Implementation Details.} The Sparse 3D VQ-VAE is configured with an $8 \times 8 \times 8$ latent grid, where each token maintains a feature dimension of 64. The codebook $\mathcal{C}$ is comprised of 4,096 entries, with the 0-th index specifically reserved as a zero token to represent unoccupied space. Model weights are initialized from the TRELLIS~\cite{xiang2025structured} VAE, followed by a two-stage training procedure consisting of 60,000 steps per stage. This training is conducted using 8 NVIDIA A100 GPUs. For the core reasoning engine, we utilize the Qwen3-VL-8B architecture as the backbone MLLM, which undergoes fine-tuning for 30,000 steps on a cluster of 32 NVIDIA A100 GPUs. A structured prompt template is employed during the training phase to effectively unify multimodal inputs. Comprehensive details regarding the prompt structures and representative QA pair examples are documented in the Appendix.

\begin{figure*}[t]
  \centering
  \includegraphics[width=\linewidth]{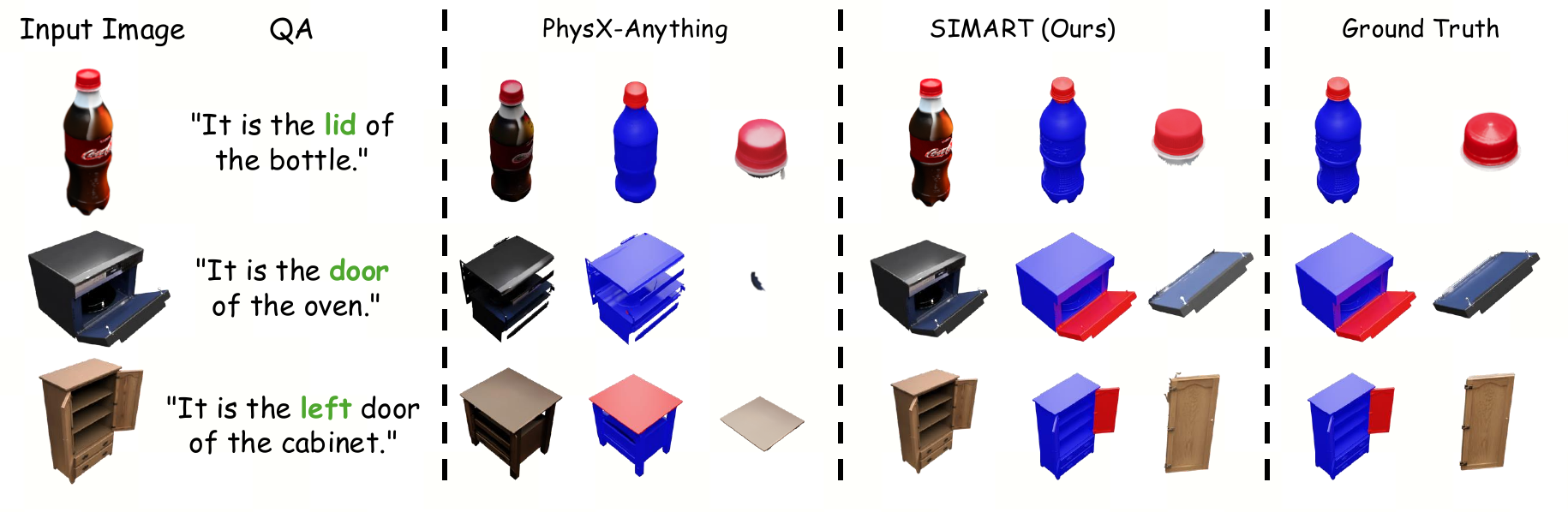}
  \caption{Qualitative comparison of part grounding capabilities under descriptions for AI-generated objects. The results demonstrate that SIMART precisely identifies and isolates functional components such as lids and doors, maintaining superior geometric consistency with the ground truth.}
  \label{fig:gd}
\end{figure*}

\subsection{Articulated Object and Kinematic Awareness}

\label{sec:articulation_results}

\noindent \textbf{Metrics.} To evaluate the precision of the generated articulated assets, we utilize five quantitative indicators. The correctness of joint classification is measured by Type Accuracy ($\text{Type} \uparrow$). The precision of the kinematic structure is assessed through Axis Error ($\text{Axis} \downarrow$), representing the angular deviation of the predicted joint axis, and Origin Error ($\text{Origin} \downarrow$), which calculates the L2 distance between the predicted and ground-truth joint origins. Geometric decomposition quality is evaluated using Intersection over Union ($\text{IoU} \uparrow$) to measure the overlap between predicted and true part segments, and Chamfer Distance ($\text{CD} \downarrow$) to quantify the reconstruction error of the individual part meshes.

\noindent \textbf{Results and Analysis.} 
As demonstrated in \Cref{tab:method_comparison}, SIMART achieves state-of-the-art performance across all metrics on both ID and AI-generated benchmarks. We compare our approach against existing baselines including Urdformer~\cite{chen2024urdformer}, Articulate-Anything~\cite{le2024articulate}, Physx-Anything~\cite{cao2025physx-a} and Particulate~\cite{li2025particulate}.
With the exception of Particulate, existing approaches lack the capability to process raw mesh inputs and exhibit poor geometric alignment with the source data, resulting in exceptionally low IoU and high CD. 
Conversely, by leveraging the high-level reasoning capabilities of an MLLM, SIMART significantly outperforms Particulate, which relies on a standalone point segmentation module for part-level decomposition.
We present a qualitative comparison between SIMART and several state-of-the-art baselines in \Cref{fig:m_cp}. 
As observed, generative baselines such as Articulate-Anything and PhysX-Anything frequently produce overly simplified or misaligned geometries that fail to match the input observations. In contrast, SIMART achieves superior structural fidelity by directly processing the input mesh through the Sparse VQ-VAE and leveraging MLLM-driven part segmentation. 

\subsection{3D Part Understanding}
\label{sec:part_understanding}

\noindent \textbf{Metrics.} 
The 3D Part Understanding task evaluates the model's capacity to perform precise semantic-to-geometric grounding by identifying and reconstructing a specific object component based on a functional natural language description. 
Performance is quantified using two primary metrics: IoU to assess the spatial overlap between the predicted and ground-truth parts, and CD to measure the geometric fidelity of the reconstructed part surface.

\noindent \textbf{Results and Analysis.}
As illustrated by the performance disparity in \Cref{tab:grounding_ood}, SIMART significantly outperforms Physx-Anything on AI-generated items across both IoU and CD metrics. 
We also implement a baseline consisting of P3SAM integrated with Qwen3-VL-235B. In this setup, we utilize the VLM to verify whether the parts segmented by P3SAM align with the functional descriptions of grounding task.
This suggests that while generative baselines struggle with precise spatial localization on novel geometries, SIMART effectively leverages coordinate-aware tokenization and the extensive world knowledge of the VLM backbone to link functional descriptions to physical coordinates. 

\begin{table}[t]
  \centering
  \caption{Quantitative comparison of part grounding performance on AI-generated items. The results demonstrate the superior ability of SIMART to precisely localize and reconstruct functional components within novel geometric structures compared to generative baselines.}
  \label{tab:grounding_ood}
  \setlength{\tabcolsep}{12pt}
  \begin{tabular}{lcc}
    \toprule
    \multirow{2}{*}{Method} & \multicolumn{2}{c}{AI-generated Items} \\
    \cmidrule(lr){2-3}
    & IOU $\uparrow$ & CD $\downarrow$ \\
    \midrule
    Physx-Anything~\cite{cao2025physx-a}      & 0.067 & 0.347 \\
    P3SAM~\cite{ma2025p3} + Qwen3-VL      & 0.507 & 0.234 \\
    \textbf{SIMART (ours)} & \textbf{0.807} & \textbf{0.018} \\
    \bottomrule
  \end{tabular}
\end{table}

\subsection{Ablation Studies}
\label{sec:ablation}

The ablation study systematically evaluates the impact of sequence length reduction, the sparse zero token mechanism, and the integration of visual features on the framework's overall performance.
Notably, the MLLM is required to output a complete latent voxel grid to the decoder to facilitate the reconstruction of each segmented part. 
The dense baseline necessitates the generation of the entire voxel grid per component. Consequently, the total sequence length scales linearly with the number of parts, frequently surpassing the memory capacity of the Qwen3-VL backbone during the fine-tuning phase.
This scaling behavior precipitates Out-of-Memory (OOM) errors, thereby precluding successful training for complex, multi-part objects.
To mitigate the sequence length bottleneck, we evaluated a force sparse configuration. 
Leveraging the principle of sparsity, this approach retains tokens only at occupied voxel coordinates, resulting in a significant reduction in sequence length. 
By leveraging the specialized zero-token mechanism, our framework achieves superior performance across all evaluation metrics while utilizing a minimal number of tokens.
The final integration of visual features to form the complete SIMART model yields the highest performance across all evaluation metrics. 
This highlights the critical role of visual information in resolving geometric ambiguities, particularly in cases where objects share similar morphologies but possess distinct articulation structures.
The token counts reported in \Cref{tab:ablation} represent the average number of tokens across the entire training dataset, which contains objects with varying part counts (averaging four parts per object). 
The reported counts encompass the part-specific latent voxel tokens and the symbolic text tokens generated exclusively for the output sequence.

\begin{table}[t]
  \centering
  \caption{Ablation study evaluating different component of SIMART on AI-generated items across kinematic and geometric performance metrics.}
  \label{tab:ablation}
  \setlength{\tabcolsep}{3pt}
  \begin{tabular}{lccccc}
    \toprule
    Method & Type $\uparrow$ & Center $\downarrow$ & IoU $\uparrow$ & CD $\downarrow$ & Token Num $\downarrow$ \\
    \midrule
    + Dense token            & \multicolumn{4}{c}{OOM} & 4138 \\
    + Force Sparse           & 0.661 & 0.157 & 0.678 & 0.100 & 862 \\
    + Zero Sparse            & 0.794 & 0.108 & 0.745 & 0.074 & \textbf{516} \\
    + Vision (\textbf{ours}) & \textbf{0.937} & \textbf{0.074} & \textbf{0.832} & \textbf{0.055} & \textbf{516} \\
    \bottomrule
  \end{tabular}
\end{table}

\begin{figure}[t] 
  \centering
  \begin{subfigure}[b]{0.5\linewidth}
    \centering
    \includegraphics[width=\linewidth, height=5cm, keepaspectratio]{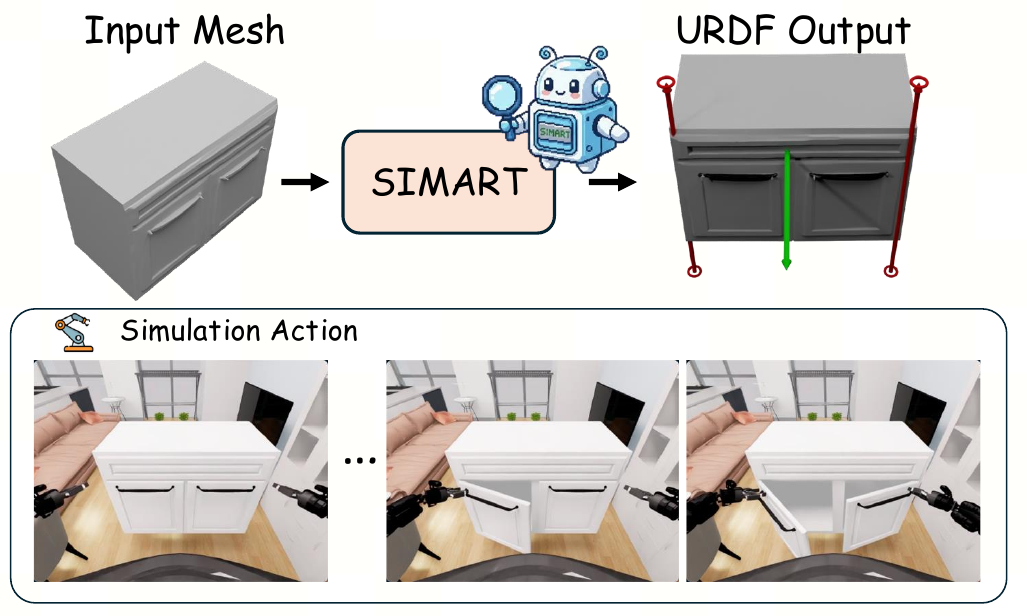}
    \caption{Deployment in physics-based simulator.}
    \label{fig:physics_sim}
  \end{subfigure}
  \begin{subfigure}[b]{0.4\linewidth}
    \centering
    \includegraphics[width=\linewidth, height=6cm, keepaspectratio]{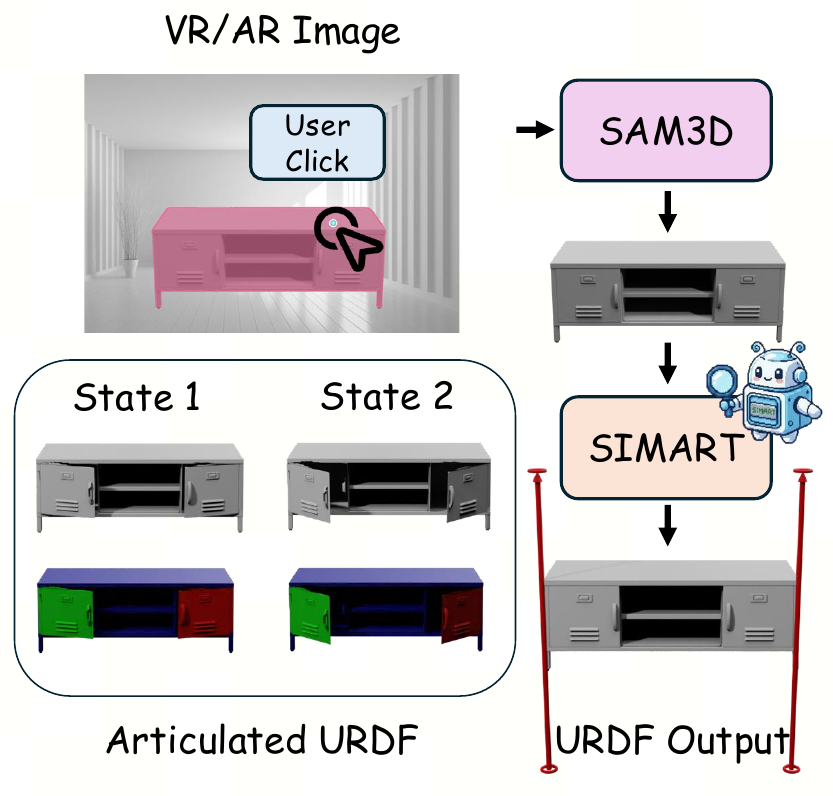}
    \caption{Interactive VR/AR generation.}
    \label{fig:vrar_app}
  \end{subfigure}

  \caption{Applications of SIMART. \textbf{Left}: The framework converts static images into articulated models for robotic manipulation. \textbf{Right}: Integration with SAM3D~\cite{sam3dteam2025sam3d3dfyimages} enables click-to-functionalize asset generation.}
  \label{fig:combined_apps}
\end{figure}

\section{APPLICATIONS}

\subsection{Physics-based Simulation}
\label{sec:physics_sim}

\Cref{fig:physics_sim} showcases the capability of SIMART to transform raw mesh into high-fidelity assets optimized for interactive environments. By integrating reconstructed geometries with predicted kinematic structures, these assets can be directly imported into simulators like NVIDIA Isaac Sim~\cite{makoviychuk2021isaac} for rigorous robotic manipulation testing. We utilize the inherent reasoning of the MLLM to estimate real-world scales, ensuring physical consistency within the virtual space. The execution of complex physical interactions underscores the superior structural and functional quality of the articulated assets produced by our method. We strongly recommend readers to check out our supplementary video results.
This automated pipeline provides three primary advantages for embodied AI: the scalable generation of diverse training scenarios, the facilitation of interactive learning via real-time dynamic feedback, and the provision of multi-modal observation data to benchmark advanced vision-language-action (VLA) models.

\subsection{VR/AR}
\label{sec:vrar}

Beyond robotic simulation, SIMART extends to immersive VR/AR environments by facilitating user-driven interactive asset creation as shown in~\Cref{fig:vrar_app}. Through a simple click-based selection interface, the system integrates geometric generation from SAM3D~\cite{sam3dteam2025sam3d3dfyimages} with SIMART to generate articulated digital twins. This workflow allows users to transform static virtual surroundings into interactive components with realistic kinematic constraints. By functionalizing captured high-fidelity meshes into simulation-ready assets, our framework significantly enriches the interaction logic of virtual worlds, enabling seamless manipulation in mixed-reality scenarios.

\section{Conclusion}
\label{sec:conclusion}

In this paper, we introduced SIMART, a multimodal framework that transforms static 3D meshes into functional, simulation-ready assets by decoupling kinematic reasoning from geometric generation. By leveraging a Sparse 3D VQ-VAE, our approach achieves a 70\% reduction in token redundancy, effectively resolving the memory-exhaustion issues inherent in dense volumetric representations. Our method utilizes the Qwen3-VL backbone to perform precise part decomposition and joint parameter estimation across diverse object categories. We also proposed SIMART-Bench, a high-fidelity benchmark that establishes a standardized metric for evaluating articulation accuracy on both in-domain and out-of-distribution assets. Experimental results demonstrate that SIMART significantly outperforms generative baselines by maintaining strict structural fidelity to the original input geometry. 

Despite these advancements, the scarcity and inconsistent quality of existing articulated datasets remain a primary limitation for open-world generalization. Future work will focus on utilizing SIMART as a foundational tool to generate pre-verified articulation predictions, thereby accelerating the data-annotation loop. 
This automated pipeline facilitates the creation of larger and more diverse datasets, thereby further enhancing the generative capabilities for synthesizing simulation-ready articulated objects.

\clearpage

\bibliographystyle{plainnat}
\bibliography{main}

\clearpage

\beginappendix

\section{Demo Video Details}
We provide a supplementary demo video to better showcase our results and the potential applications for downstream robot manipulation tasks. 
We construct a simulation environment using background from~\cite{jin2025artvip} and integrate our synthesized URDF assets for interactive manipulation. 
Furthermore, the video demonstrates a diverse range of kinematic motion sequences and articulated behaviors across various object categories.

\section{Sparse 3D VQ-VAE Implementation Details}

The architectural design of our Sparse 3D VQ-VAE is optimized to balance reconstruction fidelity with the sequence length constraints of the MLLM backbone. 
We initially encode the $64 \times 64 \times 64$ voxel grid into a $16 \times 16 \times 16$ latent grid to capture fundamental geometric structures. 
To further compress the representation for efficient multimodal reasoning, we aggregate every eight neighboring tokens along the channel dimension, resulting in an $8 \times 8 \times 8$ latent grid with a feature dimension of 64 per token. 
While conventional 3D generative models often require larger codebooks, the high summarizing capability of our specialized zero-token mechanism allows us to reduce the codebook size to 4,096 entries. 
This reduction significantly lowers the computational overhead while maintaining high-fidelity reconstruction of manifold surfaces. 

Following standard VQ-VAE training protocols, we employ a binary cross-entropy reconstruction loss $\mathcal{L}_{\text{rec}}$ and a commitment loss to stabilize codebook learning, formulated as:
\begin{equation}
    \begin{aligned}
        \mathcal{L}_{total} = \mathcal{L}_{rec}(G_{geo}, \hat{G}_{geo}) &+ \|sg[E(G_{geo})] - \hat{z}\|_2^2 \\
        &+ \beta\|E(G_{geo}) - sg[\hat{z}]\|_2^2
    \end{aligned}
\end{equation}
where $sg[\cdot]$ denotes the stop-gradient operator and $\beta$ is a weighting hyperparameter. 
Furthermore, the Sparse 3D VQ-VAE is pre-trained on a 500k-object subset following the TRELLIS~\cite{xiang2025structured} data distribution to ensure high-fidelity geometric reconstruction.

To rigorously assess the geometric fidelity of our Sparse 3D VQ-VAE, we evaluate the reconstruction error by comparing the original input 3D mesh with the mesh generated from the decoded voxel grid. 
The input meshes are first voxelized into a $64 \times 64 \times 64$ grid to serve as the ground truth. 
After the encoding, quantization, and decoding processes, the resulting occupancy grid is compared against the original input to compute the reconstruction metrics. We employ two primary indicators: Mean Squared Error (MSE) and Chamfer Distance (CD), both calculated at the $64 \times 64 \times 64$ voxel resolution. These metrics quantify the pixel-wise occupancy alignment and the overall surface deviation, respectively, providing a clear measure of how well the sparse tokens preserve the underlying manifold geometry of the articulated parts.

\begin{table}[htbp]
  \centering
  \caption{Ablation study of Sparse 3D VQ-VAE configurations regarding reconstruction quality. For enhanced readability, all reported MSE and CD values are scaled by $10^{5}$.}
  \label{tab:vqvae_detail_ablation}
  \setlength{\tabcolsep}{12pt}
  \begin{tabular}{lcc}
    \toprule
    Configuration & MSE $\downarrow$ & CD $\downarrow$ \\
    \midrule
    Sparse 8*8*8 (Ours)       & 1.84 & 4.19 \\
    Sparse $16 \times 8 \times 8$ & 1.15 & 2.27 \\
    Codebook (8192)           & 1.84 & 4.56 \\
    Force Sparse (No Zero Token) & 2.66 & 56.10 \\
    \bottomrule
  \end{tabular}
\end{table}

\noindent \textbf{Analysis of Sparse Reconstruction Quality.} The results in \Cref{tab:vqvae_detail_ablation} demonstrate that the integration of the zero token mechanism significantly enhances the reconstruction capabilities of the Sparse VQ-VAE, particularly in reducing the Chamfer Distance compared to the force sparse baseline lacking this feature. While expanding the latent resolution to $16 \times 8 \times 8$ provides further gains in geometric fidelity, such a configuration doubles the resulting token sequence length, which introduces a substantial memory overhead for the multimodal backbone. As a critical trade-off to ensure efficient fine-tuning without memory exhaustion, the $8 \times 8 \times 8$ latent grid is implemented to maintain a compact token representation while preserving the structural details necessary for robot operation learning. Furthermore, doubling the codebook size to 8,192 does not yield significant improvements in the reconstruction metrics, indicating that a 4,096-entry codebook is sufficient when utilizing the zero token to summarize unoccupied regions. This optimized architecture allows SIMART to capture complex 3D manifolds effectively within the sequence length constraints of the MLLM. The balanced performance of this sparse representation confirms its suitability for generating functional, simulation-ready assets in real-to-sim pipelines.

\noindent \textbf{Analysis of Emergent Zero Tokens.} The implementation of our Force Sparse mechanism is grounded in the observation that VQ-VAE models naturally develop specialized codebook entries to represent unoccupied space. During our training of a dense VQ-VAE baseline, we observed that even without explicit zero-token supervision, approximately two to four codebook entries consistently converge to represent the null distribution of empty voxels. These entries, when processed by the decoder, effectively reconstruct empty volumetric regions with high stability. For example, in our experimental dense model, the entry indexed as \textbf{voxel-1849} was identified as a functional surrogate for empty space. By formalizing this behavior and explicitly reserving the 0-th index as a dedicated zero token, we achieve a more robust and interpretable sparse representation. This strategy prevents the MLLM from wasting attention on irrelevant background tokens, thereby focusing its reasoning capacity on the functional and articulated parts of the 3D asset.

\section{MLLM Implementation Details}
\label{sec:mllm_details}

The SIMART framework utilizes Qwen3-VL-8B as the foundational multimodal backbone for all experiments. This model is integrated into a comprehensive inference pipeline that processes a concatenated sequence comprising the system prompt, sparse voxel tokens, task-specific questions, and visual inputs. The input mesh representation follows a coordinate-aware format structured as <voxel> xyz <mesh-token>, where xyz denotes the quantized spatial coordinates and mesh-token represents the discrete latent feature from the Sparse 3D VQ-VAE codebook. To resolve scale ambiguities and provide global context, each object is rendered as a 252x252 pixel image from a 45-degree isometric perspective.

The model handles two primary tasks through specialized question templates. For kinematic reasoning and URDF asset generation, the model is queried with the instruction: \textbf{Describe the object with real scale and separate the object to different functional parts with each physical properties.} For the part grounding task, the model utilizes the instruction: \textbf{Generate the part of this object with description: \textit{text}}, where the \textbf{\textit{text}} variable is populated with a semantic description of a functional component.

The operational constraints and output formatting requirements are defined by a high-level system prompt, which is essential for maintaining structural consistency and adhering to physical logic for real-to-sim transfer. The complete system prompt used for training and inference is detailed in \Cref{tab:system_prompt}.

\begin{table*}[htbp]
  \centering
  \caption{Detailed system prompt for the SIMART multimodal reasoning and URDF generation tasks.}
  \label{tab:system_prompt}
  \begin{small}
  \begin{tabular}{|p{0.95\textwidth}|}
    \hline
    \vspace{2pt} \\
    \texttt{You are a multimodal 3D reasoning assistant operating on SPARSE VOXEL REPRESENTATIONS.} \\
    \vspace{5pt} \\
    \textbf{\#\# VOXEL REPRESENTATION} \\
    The 3D space is represented as a $16 \times 8 \times 8$ voxel grid. Each occupied voxel is encoded as three atomic tokens in the exact order: \texttt{<voxel> [xyz] [K]}. \\
    \textbf{Coordinate System (right-handed):} \\
    - x: left $\to$ right (0--15); y: front $\to$ back (0--7); z: ground $\to$ top (0--7). \\
    \textbf{Index mapping:} $xyz = 64x + 8y + z$ (range: 0--1023). \\
    \textbf{K:} An atomic geometry token ($K \in [0, 8191]$) representing local surface geometry. \\
    
    \vspace{5pt} \\
    \textbf{\#\# TASK: OBJECT2URDF} \\
    \textbf{Input:} A sequence of tokens representing the sparse voxel structure of a 3D object. \\
    \textbf{Goal:} Decompose the object into functional parts, predict real-world scale, and specify kinematic parameters (fixed, revolute, prismatic, free, hinge, rigid). \\
    \textbf{Output (Strict JSON only):} \texttt{\{"object\_captions": \{...\}, "parts\_captions": \{...\}, "parts\_voxels": \{...\}\}} \\
    
    \vspace{5pt} \\
    \textbf{\#\# TASK: OBJECT PART REASONING AND LOCALIZATION} \\
    \textbf{Input:} Voxel sequence and a functional question describing a specific part. \\
    \textbf{Goal:} Identify and localize the specific part; output the voxel tokens corresponding to that part. \\
    
    \vspace{5pt} \\
    \textbf{\#\# STRICT RULES} \\
    - \textbf{Output voxels} must be a subset of the input. \\
    - \textbf{center}: $[x, y, z]$ integers in $[0, 200]$, representing grid points at 0.005 resolution. \\
    - \textbf{axis}: $[dx, dy, dz]$ integers in $[0, 100]$, representing a direction vector. \\
    - \textbf{limits}: $[-val, val]$ where $100$ represents $180^{\circ}$ for revolute or max distance for prismatic. \\
    - \textbf{No extra text, headers, or formatting beyond the JSON.} \\
    \vspace{2pt} \\ \hline
  \end{tabular}
  \end{small}
\end{table*}

The SIMART framework generates structured outputs designed for seamless integration into downstream robotics pipelines. 
First, it outputs discrete voxel tokens for each functional component, which are subsequently reconstructed into the 3D space via the sparse 3D VQ-VAE decoder to form the part-segmented meshes $\mathcal{M}_{seg}$. 
Concurrently, the model generates a structured JSON-like representation $T_{desc}$.
This description constitutes the simulation metadata $\mathcal{P}_{sim}$, explicitly specifying physical attributes such as material type and density, as well as the hierarchical kinematic tree including joint origins, axes, and limits. 
Crucially, the geometry for each part is represented as a discrete sequence of sparse <mesh-tokens>, ensuring precise part-level decomposition. In grounding tasks, the model outputs the specific <mesh-tokens> that correspond to the functional part described in the query. The standardized formats for these outputs are summarized in \Cref{tab:model_output_example_single}.

\begin{table}[htbp]
  \centering
  \caption{Representative output generated by SIMART for a URDF generation task.}
  \label{tab:model_output_example_single}
  \begin{small}
  \begin{tabular}{|p{0.96\columnwidth}|}
    \hline
    \vspace{2pt} \\
    \texttt{\{} \\
    \texttt{\ \ "object\_captions": \{} \\
    \texttt{\ \ \ \ "name": "Storage Box with Frame",} \\
    \texttt{\ \ \ \ "scale": 40.0} \\
    \texttt{\ \ \},} \\
    \texttt{\ \ "parts\_captions": \{} \\
    \texttt{\ \ \ \ "0": \{} \\
    \texttt{\ \ \ \ \ \ "type": "fixed",} \\
    \texttt{\ \ \ \ \ \ "material": "Plastic",} \\
    \texttt{\ \ \ \ \ \ "density": "1.2 g/cm\^{}3",} \\
    \texttt{\ \ \ \ \ \ "Young's Modulus (GPa)": 2.5} \\
    \texttt{\ \ \ \ \},} \\
    \texttt{\ \ \ \ "1": \{} \\
    \texttt{\ \ \ \ \ \ "type": "revolute",} \\
    \texttt{\ \ \ \ \ \ "parent": "0",} \\
    \texttt{\ \ \ \ \ \ "center": [100, 138, 101],} \\
    \texttt{\ \ \ \ \ \ "axis": [100, 0, 0],} \\
    \texttt{\ \ \ \ \ \ "limits": [-54, 45]} \\
    \texttt{\ \ \ \ \}} \\
    \texttt{\ \ \},} \\
    \texttt{\ \ "parts\_voxels": \{} \\
    \texttt{\ \ \ \ "0": "<voxel> 0 1785 <voxel> 1 649 ...",} \\
    \texttt{\ \ \ \ "1": "<voxel> 43 1930 <voxel> 44 13 ..."} \\
    \texttt{\ \ \}} \\
    \texttt{\}} \\
    \vspace{2pt} \\ \hline
  \end{tabular}
  \end{small}
\end{table}

\section{Generation Asset Benchmark Build Pipeline}
\label{sec:benchmark_pipeline}
We constructed the \textbf{SIMART-Bench} dataset, comprising diverse AI-generated objects, to address the limitations of existing evaluation protocols. Most current methods are tested predominantly on PartNet-Mobility models, which may obscure the true generative capabilities and generalizability of the proposed architectures. As 3D asset generation has advanced rapidly, there is now an abundance of high-quality, AI-generated raw meshes that lack functional articulation. Demonstrating the capacity to process these diverse, unstructured generated objects serves as a robust validation of the practical applicability and real-world deployment potential of our proposed framework.
SIMART-Bench currently consists of over 10 categories of articulated objects, comprising 36 unified assets specifically curated for comprehensive evaluation.

To establish a high-fidelity evaluation standard for our articulated asset generation, we developed a systematic annotation pipeline to create ground-truth (GT) labels for the generated models. The process begins with an automated segmentation phase utilizing the P3SAM method to decompose the raw 3D assets into initial segments. Empirical observations indicate that this automated step frequently results in severe over-segmentation, typically partitioning a single object into 6 to 10 distinct fragments. To address this, we implement a human-in-the-loop refinement stage where these segments are manually merged to consolidate over-segmented parts into 2 to 4 functional components. This refinement ensures that the segmented parts correspond strictly to the movable entities required for realistic kinematic simulation.

Following the structural decomposition, we utilize a specialized Web UI to perform precise annotation of the motion axes and joint positions for each articulated component. This manual labeling process defines the kinematic hierarchy and physical constraints necessary for URDF generation. By combining automated segmentation with expert-guided merging and annotation, our pipeline generates accurate, simulation-ready ground-truth metadata. This high-quality data serves as the foundation for our benchmark, enabling the rigorous evaluation of articulation accuracy and geometric fidelity across diverse object categories in our real-to-sim framework.

\begin{figure*}[p]
  \centering
  \includegraphics[width=0.95\linewidth]{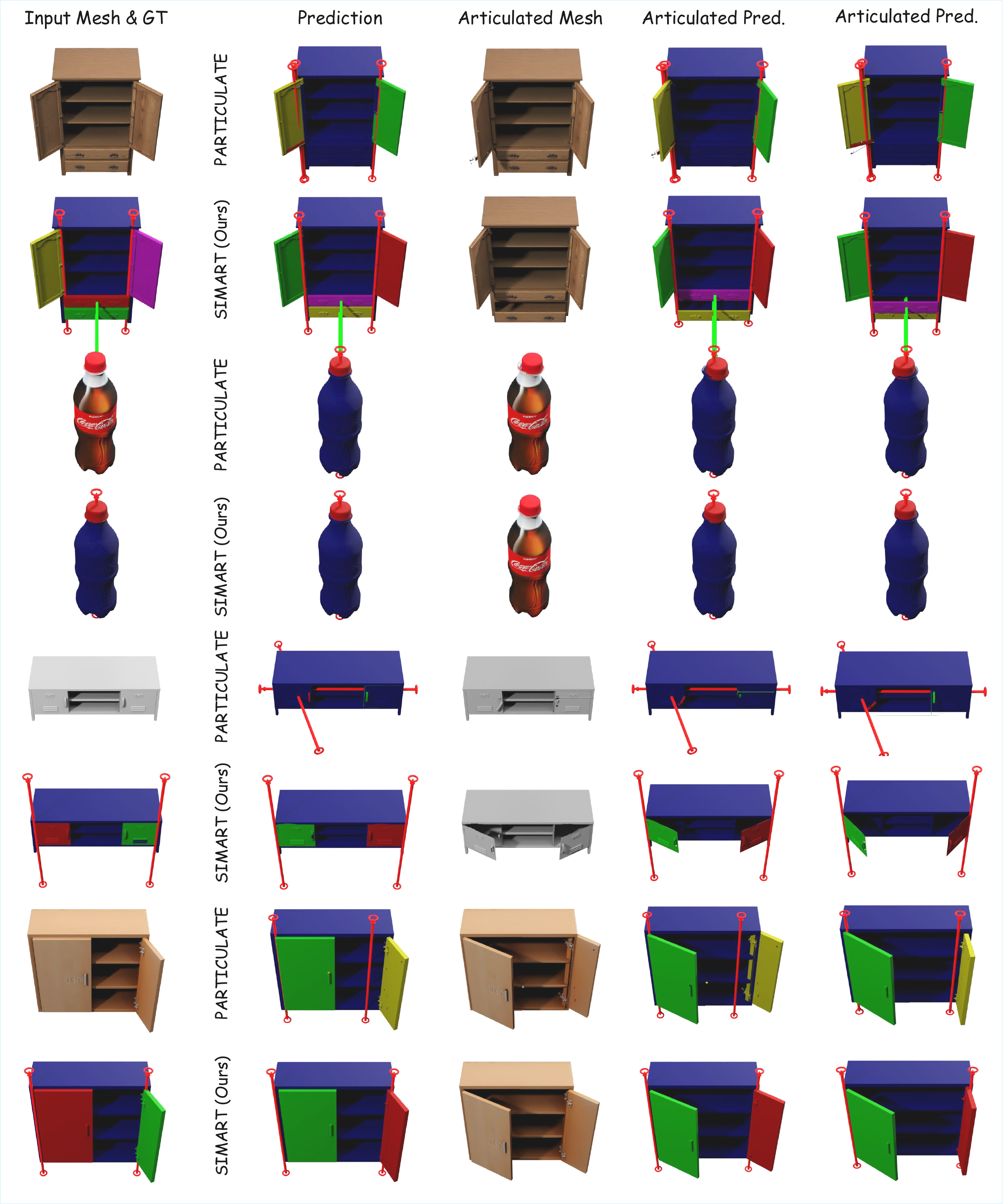}
  \caption{Extended qualitative comparison between SIMART and the Particulate baseline. All displayed samples are AI-generated objects.}
  \label{fig:comparison_extended}
\end{figure*}


\begin{figure*}[p]
  \centering
  \includegraphics[width=0.96\linewidth]{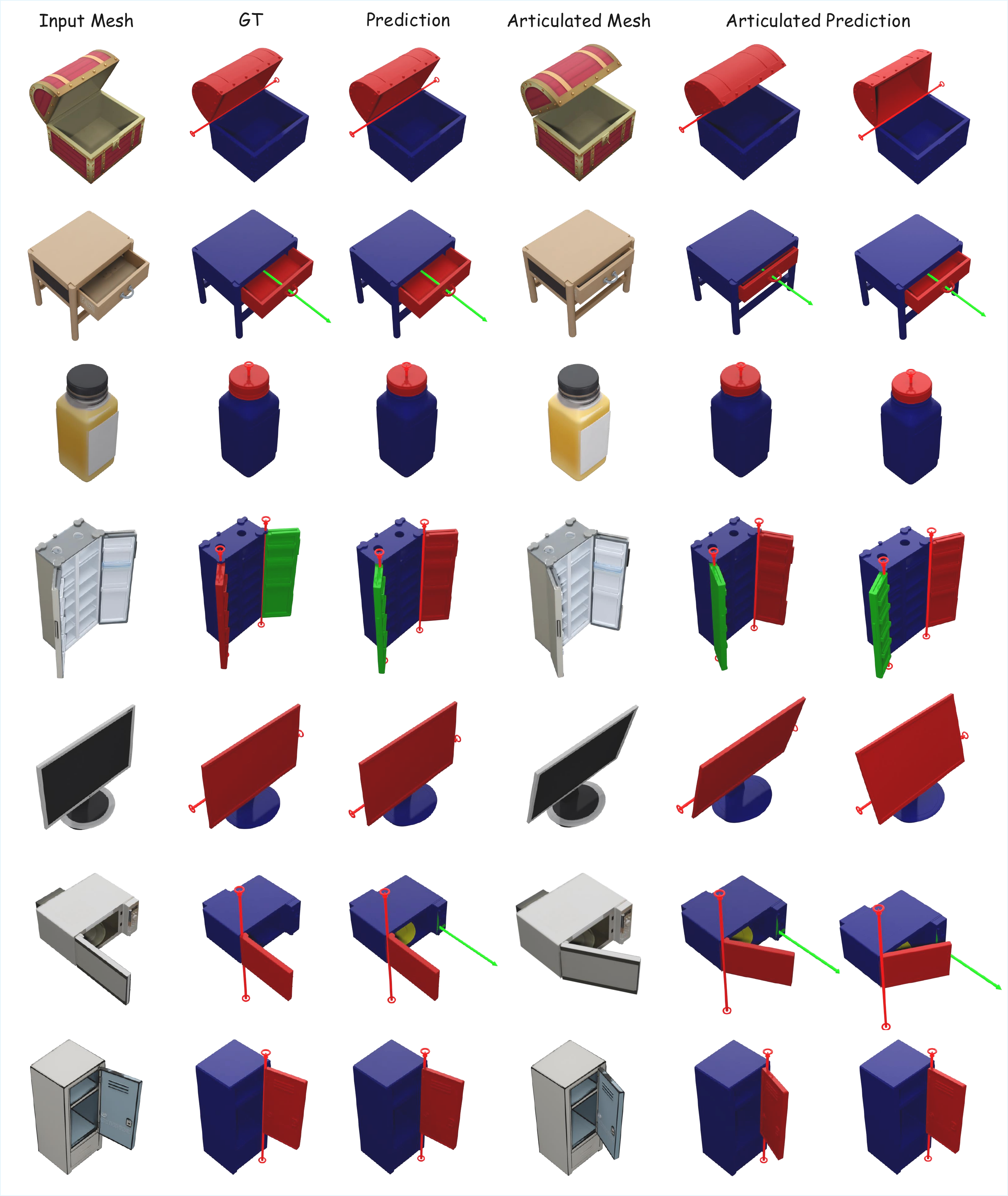}
  \caption{Comprehensive gallery of simulation-ready assets generated by the SIMART framework. Every asset shown is an AI-generated objects.}
  \label{fig:results_gallery}
\end{figure*}

\end{document}